\documentclass{article}

\usepackage{arxiv}

\usepackage[utf8]{inputenc} 
\usepackage[T1]{fontenc}    
\usepackage{hyperref}       
\usepackage{amsfonts}       
\usepackage{nicefrac}       
\usepackage{microtype}      
\usepackage{lipsum}
\usepackage{graphicx}
\graphicspath{ {./images/} }
\usepackage{hyperref}
\usepackage{url}
\usepackage{multirow}      
\usepackage{booktabs}
\usepackage{graphicx}
\usepackage{amsmath}
\usepackage{colortbl}    
\usepackage[table]{xcolor} 
\usepackage{makecell}
\usepackage{subcaption}
\usepackage{array} 
\usepackage{adjustbox} 
\usepackage{changepage}
\usepackage{wrapfig} 
\usepackage{caption}
\usepackage{parskip}

\usepackage{enumitem}

\definecolor{darkblue}{RGB}{102,200,255}
\definecolor{lightblue}{RGB}{204,229,255}

\title{MSCoRe: A Benchmark for Multi-Stage Collaborative Reasoning in LLM Agents}

\author{
 Yuzhen Lei \\
  Jilin University\\
  \texttt{leiyz9921@mails.jlu.edu.cn} \\
   \And
 Hongbin Xie \\
 Southern University of Science and Technology\\
  \texttt{12131108@mail.sustech.edu.cn} \\
  \And
  Jiaxing Zhao \\
  Jilin University\\
  \texttt{zhaojx9921@mails.jlu.edu.cn} \\
   \And
  Shuangxue Liu \\
  Jilin University\\
  \texttt{liusx9921@mails.jlu.edu.cn} \\
   \And
   Xuan Song\footnote{Corresponding author: songxuan@jlu.edu.cn}\\
  Jilin University\\
  \texttt{songxuan@jlu.edu.cn} \\
}

\begin{document}
\maketitle
\begin{abstract}
Large Language Models (LLMs) have excelled in question-answering (QA) tasks within single domains. However, their reasoning and coordination capabilities in complex, multi-stage scenarios remain underexplored. Existing benchmarks typically focus on isolated tasks or narrow domains, overlooking models' abilities for multi-stage collaboration and optimization without explicit external guidance. To bridge this gap, we propose \textbf{MSCoRe}, a novel benchmark comprising 126696 domain-specific QA instances spanning scenarios in automotive, pharmaceutical, electronics, and energy sectors. The dataset is created using a structured three-phase pipeline: dynamic sampling, iterative question-answer generation, and a multi-level quality assessment to ensure data quality. Tasks are further categorized into three difficulty levels according to stage coverage and complexity.  With MSCoRe, we have conducted a comprehensive evaluation of various state-of-the-art LLM agents. The commercial models performed best across all tasks and scenarios, but a notable gap in ROUGE scores remains between simple and complex tasks. We also tested the models' robustness and found that their performance is negatively affected by noisy data. MSCoRe provides a valuable new resource for the community to evaluate and improve multi-stage reasoning in LLM agents. The code and data are available at \url{https://github.com/D3E0-source/MSCoRE}.
\end{abstract}

\section{Introduction}
\setlength{\parindent}{0pt} 
\begin{wrapfigure}{r}{0.4\textwidth}  
    \vspace{-1.2em}  
    \centering
    \includegraphics[width=0.4\textwidth,trim=10 0 0 0, clip]{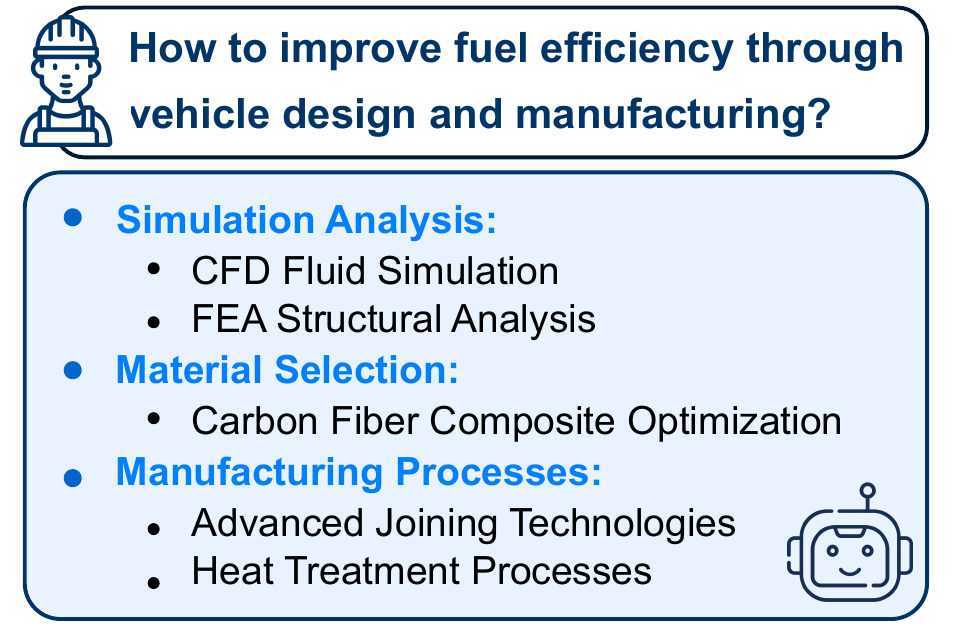}
    \caption{An illustrative multi-stage QA example highlighting design–manufacturing synergy for vehicle fuel efficiency improvement in MSCoRe}
    \label{Figure_1}
    \vspace{-\baselineskip}  
\end{wrapfigure}
Open-domain question answering (QA) serves as a cornerstone task for evaluating advanced capabilities of Artificial Intelligence(AI) language models \cite{10.1162/tacl_a_00276}. It requires both precise intent understanding and the generation of coherent, logically sound responses. A rich ecosystem of benchmarks has been developed to this end. General-purpose benchmarks such as WikiQA \cite{yang-etal-2015-wikiqa}, MMLU \cite {hendrycks2020measuring}, CommonsenseQA \cite{talmor2018commonsenseqa}, and ZeroShot \cite {sun2025zerosearch} span diverse domains and provide rigorous evaluation of models’ foundational reasoning and knowledge. Currently, domain-specific datasets such as EQUALS \cite {chen2023equals} for law, DISC-FinLLM\cite{chen2023disc} for finance, and MedExQA\cite{kim2024medexqa} for medicine have emerged to assess specialized expertise. However, a critical limitation persists across these evaluations: they predominantly treat tasks as isolated, single-stage problems. This approach assesses knowledge within a specific context, but overlooks the crucial interdependencies and collaborative reasoning required across different operational stages within a complex domain. This limitation is particularly salient in real-world industrial applications where workflows are inherently multi-staged. For example, the automotive value chain involves interconnected stages from design and manufacturing to supply chain and quality inspection\cite{pinho2022developing}.  A design choice directly impacts manufacturing feasibility, while a supply chain disruption can halt production. Similarly, in the pharmaceutical industry,  quality inspection results provide essential feedback for optimizing production processes\cite{gloria2024evaluating}. Effectively navigating these environments demands more than isolated knowledge; it requires the ability to perform collaborative reasoning that considers downstream consequences and upstream constraints. Existing evaluation frameworks are ill-equipped to measure this vital, holistic problem-solving capability.

\begin{figure}[h]
\centering
    \centering
    \includegraphics[width=0.7\linewidth]{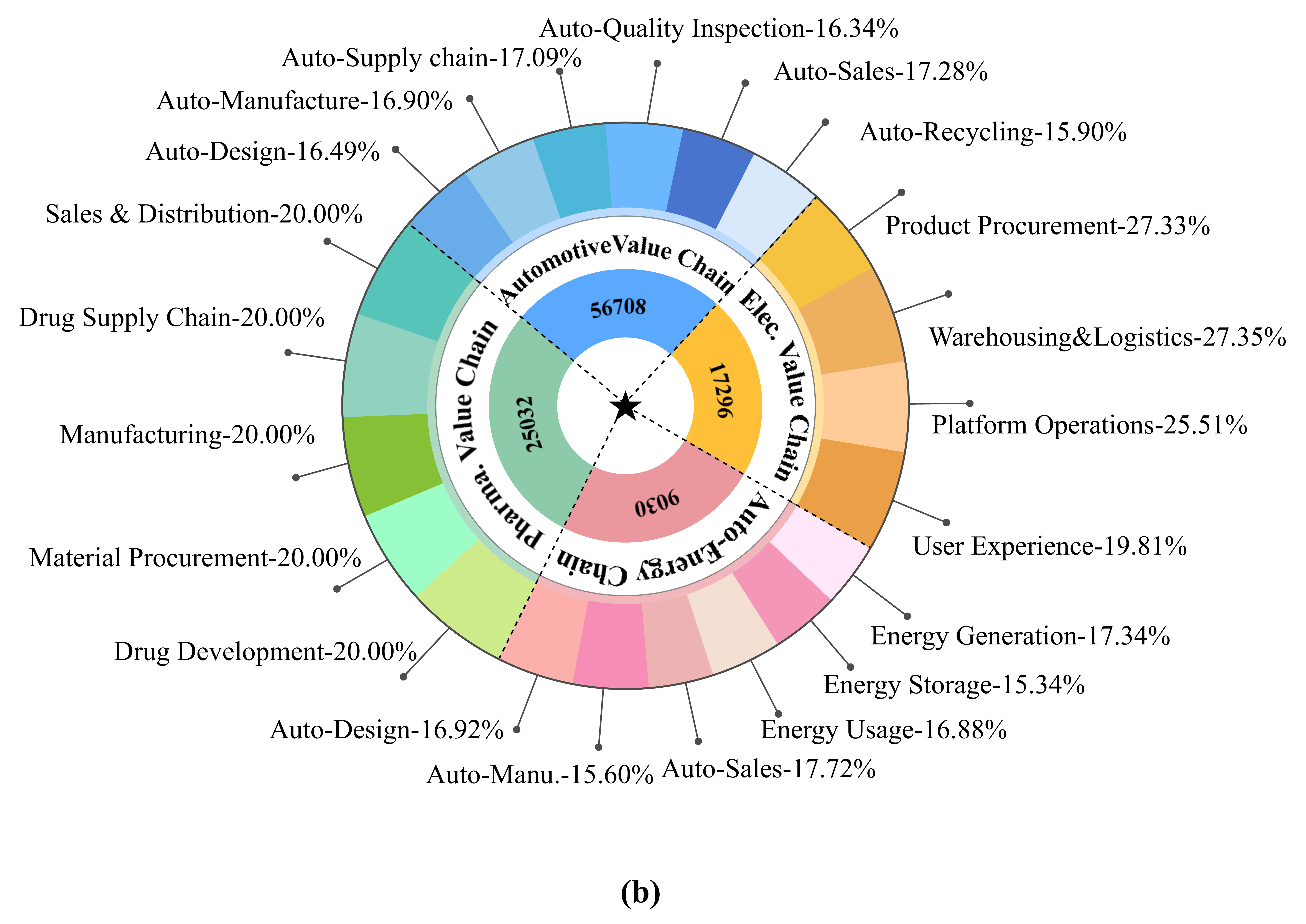}
    \caption{Data distribution across various industry sectors and workflow stages.}
    \label{figure_2_a}
\end{figure}

To bridge this crucial evaluation gap, we introduce MSCoRe, a novel benchmark designed to assess the multi-stage collaborative reasoning of LLMs. Different from existing benchmarks, MSCoRe presents complex scenarios where a viable solution requires synthesizing information and considering trade-offs across multiple, interdependent stages (as shown in Figure~\ref{Figure_1}). The benchmark comprises 1080,00 high-quality QA pairs spanning four intricate domains: automotive, pharmaceutical, e-commerce, and energy value chain(as shown in Figure~\ref{figure_2_a}). We constructed this dataset using a systematic pipeline featuring high-quality seed data, refined prompt engineering, and a multi-level quality control mechanism with feedback-driven optimization. To enable a fine-grained analysis, we stratify the tasks into three difficulty levels—simple, medium, and difficult—based on their complexity and the number of reasoning stages involved.

With MSCoRe, we have conducted a comprehensive evaluation of various state-of-the-art LLMs, including the commercial models like GPT-4o\cite{achiam2023gpt}, GPT-3.5-turbo, and Claude-3.5-haiku, as well as open-source models like Deepseek-R1\cite{guo2025deepseek}, and Qwen series\cite{bai2023qwen}. The results indicate that even models excelling in simple tasks face significant challenges when tackling full-chain tasks. Furthermore, we introduced various forms of noise—including unstandardized formats, incomplete information, and semantic inaccuracies—into selected sample datasets to assess models' robustness and interference resistance under adverse input conditions. 
In summary, this work makes the following contributions:
\begin{itemize}[leftmargin=10pt]
    \item We propose a systematic automated framework to construct MSCoRe, a large-scale benchmark with 126696 multi-stage QA instances across four industrial domains. The framework incorporates dynamic sampling, iterative generation, and multi-level quality control, and its effectiveness is validated through a Turing test in which experts misclassified over 85\% of AI-generated data as human-authored, confirming its human-level quality.
    \item We conduct comprehensive experiments on 15 state-of-the-art LLMs, establishing rigorous baselines and revealing critical findings: commercial models achieve the strongest performance but still show substantial degradation from single-stage to full-chain tasks, exhibit marked brittleness under noisy or incomplete inputs, and display mixed sensitivity to few-shot examples depending on model scale and capability.
    \item We provide a foundation for future research by releasing MSCoRe as a challenging testbed and offering methodological insights that highlight key limitations of current LLM agents, guiding the development of more robust, adaptive, and practically deployable systems for complex multi-stage reasoning tasks.
\end{itemize}

\section{Related Work}
\label{headings}
\subsection{Benchmarks for General LLM capabilities}
The rapid evolution of Large Language Models (LLMs) has been accompanied by the development of comprehensive benchmarks designed to evaluate their core capabilities. Early general benchmarks like GLUE\cite{wang2018glue}and SuperGLUE\cite{sarlin2020superglue} established a foundation for evaluating natural language understanding.
However, as LLMs grew in scale and complexity, more extensive evaluation suites became necessary. Benchmarks such as MMLU assess knowledge across a massive range of subjects, while BIG-bench\cite{srivastava2023beyond} tests for a diverse set of emergent abilities. More recent efforts like HELM\cite{liang2022holistic} advocate for a multi-metric, holistic evaluation to provide a more complete picture of model performance, covering aspects like accuracy, fairness, and efficiency. Lately, the evaluation focus has shifted towards more interactive and agent-like behaviors. For instance, AgentBench\cite{liu2023agentbench} evaluates LLMs in complex, multi-turn open-ended environments, and ToolBench\cite{qin2023toolllm} assesses their ability to use external tools to solve problems. While these benchmarks are crucial for understanding the general reasoning and task-execution abilities of LLMs, they typically consist of a collection of discrete tasks. They often do not capture the intricate, causal dependencies inherent in a single, coherent industrial workflow, where the output of one stage becomes a critical and constraining input for the next. MSCoRe is specifically designed to fill this gap by modeling these inter-stage relationships.

\subsection{Domain-Specific LLM Evaluation}
Recognizing that real-world applications require specialized knowledge, there has been a significant trend towards creating domain-specific benchmarks. In high-stakes fields like medicine, benchmarks such as MedQA\cite{jin2021disease} and MedMCQA\cite{kim2024medexqa} evaluate clinical knowledge by testing models on medical exam questions. The recent evaluation framework for Med-PaLM 2 further highlighted the need for rigorous assessment against real-world medical challenges\cite{singhal2025toward}. Similarly, the finance sector has seen the introduction of benchmarks like FinQA\cite{chen2023equals} and the more recent FinEval\cite{zhang2023fineval}, which focus on financial report analysis and economic knowledge. In the legal domain, benchmarks like LexGLUE\cite{chalkidis2021lexglue} test models on legal text processing and reasoning tasks. These domain-specific benchmarks are indispensable for validating the factual accuracy and specialized knowledge of LLMs. However, their primary focus is often on assessing a model's ability to act as a "knowledge repository" or a single-turn "domain expert." For example, they might test if a model can answer a specific legal question or interpret a financial table. They generally do not evaluate a model's capacity to reason through a multi-stage process, such as navigating the entire drug development lifecycle from research to quality control, or managing the automotive supply chain from design to manufacturing. MSCoRe addresses this by focusing not just on domain knowledge, but on reasoning across the procedural stages of that domain.

\subsection{Complex and Multi-Step Reasoning in LLM Agents}
The challenge of enabling LLMs to solve complex, multi-step problems has been a major driver of research. The introduction of Chain-of-Thought (CoT) prompting\cite{wei2022chain} was a seminal step, demonstrating that eliciting intermediate reasoning steps significantly improves performance on complex tasks. This has inspired a host of more advanced reasoning strategies. Self-Consistency\cite{wang2022self} improves robustness by sampling multiple reasoning paths and selecting the most consistent answer. More structured approaches like Tree of Thoughts (ToT)\cite{yao2023tree} and Graph of Thoughts (GoT)\cite{besta2024graph} allow models to explore and self-evaluate diverse reasoning pathways, enabling them to handle problems that require planning and exploration. These techniques provide the means for models to tackle complex problems. However, a corresponding gap exists in evaluation: many benchmarks testing these abilities rely on abstract logic puzzles, mathematical problems, or game-playing scenarios. While useful, they may not reflect the constraints and dependencies of real-world industrial processes. GAIA\cite{mialon2023gaia} has begun to address complex, multi-step tasks requiring tool use, but still maintains a generalist focus. MSCoRe complements this line of research by providing a grounded, domain-specific testbed to address these challenges. It directly evaluates the practical, multi-stage reasoning capabilities and robustness of leading LLMs in the face of increasing task complexity and data noise. The benchmark measures a model's ability to solve problems where success is determined not by isolated logic, but by a holistic understanding of the interplay between different operational stages

\section{DATA}
\subsection{DATASET OVERVIEW}

\textbf{Data Fotmulation}

Our dataset follows the Alpaca instruction-tuning format, where each instance $d_i = (I_i, X_i, O_i)$ comprises three primary fields: \textit{instruction} $(I_i)$, \textit{input} $(X_i)$, and \textit{output} $(O_i)$. The \textit{instruction} field contains the question that guides the model's response generation. The \textit{output} field provides the target response that serves as the ground truth for training and evaluation. While the standard Alpaca format includes an \textit{input} field for providing additional context, our task formulation does not utilize this component, thus $X_i = \emptyset$ for all instances in our dataset. The format without additional input allows the model to generate answers directly from instructions. This distinguishes our approach from context-dependent QA tasks and emphasizes the evaluation of reasoning based on models' parametric knowledge. 
\begin{table}[h]
\centering
\caption{Distribution of instances across industrial domains and workflow stages}
\label{tab:data_stats}
\begin{tabular}{lcccc}
\toprule
\textbf{Domain} & \textbf{Easy} & \textbf{Medium} & \textbf{Hard} &\textbf{Total} \\
\midrule
Automotive & 56708 & 6034 & 6045 & 68787\\
Pharmaceutical & 25032 & 2005 & 2004 &29041\\
Electronics & 17731 & -- & 2005 & 19736\\
Auto-Energy & -- & -- & 9132 & 9132\\
\bottomrule
\end{tabular}
\end{table}

\textbf{Data Distribution}

Our dataset encompasses four representative industrial value chains, each reflecting the distinctive characteristics and workflows of its domain. The \textbf{Automotive Value Chain} illustrates the complete lifecycle of vehicles, where design choices determine downstream manufacture, production efficiency, and sustainability; it is structured into six stages: design, manufacturing, supply chain management, quality inspection, sales, and recycling. The \textbf{Pharmaceutical Value Chain} highlights the stringent and highly regulated process of bringing drugs from research to market, including drug research and development, raw material procurement, drug manufacturing, supply chain management, and sales and distribution. The \textbf{E-Commerce value chain} emphasizes the operational dynamics of digital platforms, where efficiency in product flow and service experience are critical; it covers product procurement, warehousing and logistics, platform operations, and user experience. Finally, the \textbf{Automotive–Energy Synergy Chain} focuses on the growing integration between mobility and energy systems, capturing how vehicle-related processes interact with energy infrastructures through five interconnected stages: automotive design, manufacturing, vehicle sales and charging, energy storage, and energy generation.
The specific data volume and distribution are shown in Figure~\ref {figure_2_a}.

\textbf{Task Defination}

To enable a more systematic and fine-grained evaluation of model performance, we categorize the dataset into three difficulty levels—\textbf{simple}, \textbf{medium}, and \textbf{hard}—based on the coverage of the value chain. 

\begin{itemize}[label=\textbullet, leftmargin=8pt]
\item \textbf{Easy Tasks} focus on single-stage optimization within individual value chain components. For instance, selecting lightweight materials for vehicle parts represents a localized decision-making problem with well-defined constraints and objectives. 
\item \textbf{Medium Tasks} involve coordinating between two or more interconnected stages. An exemplar task requires coordinating vehicle design and manufacturing processes to optimize fuel efficiency, necessitating cross-functional understanding and multi-objective optimization. \item \textbf{Hard Tasks} demand holistic integration across multiple value chain stages. Tasks such as integrating design, manufacturing, supply chain, and recycling for electric vehicle (EV) optimization require comprehensive system-level reasoning and the ability to balance competing objectives across the entire product lifecycle.
\end{itemize}
This hierarchical difficulty structure ensures that our benchmark can effectively discriminate between models with varying capabilities in industrial domain understanding, from basic single-stage reasoning to complex multi-stage optimization and decision-making. Table~\ref{tab:data_stats} summarizes the distribution of instances across domains and difficulty levels.

\subsection{DATA CONSTRUCTION}
\label{others}

\begin{figure}[h]
\begin{center}
\includegraphics[width=1.0\textwidth]{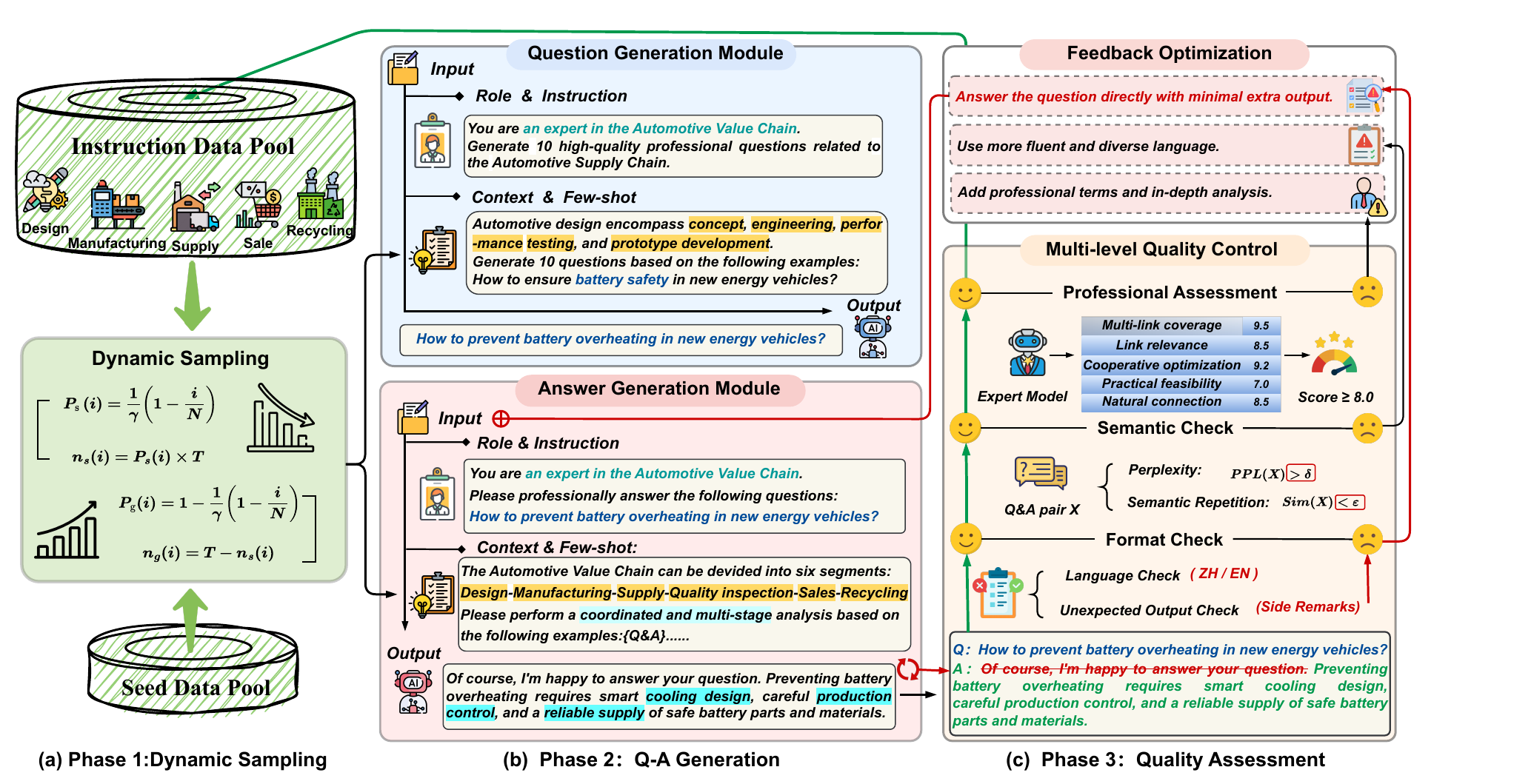} 
\end{center}
\caption{The framework of data construction: \textbf{(a)} Dynamic Sampling from seed data pool across industry domains, \textbf{(b)} Q\&A Generation with context-aware prompting and professional instruction, \textbf{(c)} Quality Assessment through multi-level evaluation including format, semantic, and expert model scoring.}
\label{framework}
\end{figure}

The construction of \textbf{MSCoRe} follows a systematic three-phase pipeline, as illustrated in Figure~\ref {framework}. This pipeline is designed to generate high-quality, diverse, and domain-specific multi-stage question-answer pairs through a closed-loop process of dynamic sampling, iterative generation, and rigorous quality assurance. Each phase is detailed below.

\textbf{PHASE 1:Dynamic Sampling Strategy}

To ensure comprehensive coverage of all stages within a domain and to promote data diversity, we employ a dynamic sampling strategy. This strategy governs the balance between leveraging our curated \textit{Seed Data Pool} and generating novel content. Instead of uniform random sampling, we define a linearly decreasing probability distribution for sampling from existing data. The sampling probability for a given topic or stage index $i$ is defined as:
\begin{equation}P_{\text{s}}^{(i)} = \frac{1}{\gamma}\left(1-\frac{i}{N}\right) 
\end{equation}
The seed samples is defined as: 
\begin{equation}
    n_{\text {s}}^{(i)}  =\left\lfloor P_{\text {s}}^{(i)} \times T\right\rfloor
\end{equation}
where $i$ denotes the topic index within a total of $N$ topics, and $\gamma$ is a hyperparameter that controls the slope of the distribution. Consequently, the probability of generating new data for topic $i$ is $P_{g} = 1 - P_{s}$. The number of generated instances is therefore $T - n_{s}$, where $K$ represents the number of examples used in generation.This strategy prioritizes sampling from well-established seed data for initial topics while progressively increasing the generation of new content for less-covered topics, thereby preventing topical imbalance and ensuring the novelty of the dataset. For a total generation target of $T$ instances per topic, the number of sampled instances n and generated instances n are determined by these probabilities.

\textbf{PHASE 2: Iterative Question-Answer Generation}

The core of our data generation relies on a two-module architecture guided by sophisticated prompt engineering, leveraging in-context learning to elicit expert-level, multi-stage reasoning from the LLM.

\begin{itemize}[label=\textbullet, leftmargin=8pt]
\item \textbf{Question Generation Module}. This module is responsible for creating high-quality, domain-specific questions. The input prompt provides the LLM with a specific Role \& Instruction, casting it as a domain expert (e.g., "You are an expert in the Automotive Value Chain") and providing a clear directive (e.g., "Generate 10 high-quality professional questions..."). To further ground the generation, the prompt also includes Context \& Few-shot examples, which provide relevant background knowledge and stylistic exemplars to guide the model.

\item \textbf{Answer Generation Module}. This module takes a generated question as input and produces a comprehensive, multi-stage answer. The prompt design is crucial here. In addition to the expert persona, the Context \& Few-shot component explicitly outlines the structure of the value chain (e.g., "Design-Manufacturing-Supply-Quality inspection-Sales-Recycling"). The model is specifically instructed to perform a "coordinated and multi-stage analysis" based on provided Q\&A exemplars that demonstrate this reasoning pattern. For instance, when answering a question about preventing battery overheating, the model is prompted to synthesize solutions from the design, manufacturing, and supply chain stages, rather than providing a siloed, single-stage answer.

\end{itemize}

\textbf{PHASE 3: Multi-level Quality Control}

To guarantee the fidelity and utility of \textbf{MSCoRe}, we implement a rigorous, multi-faceted quality assurance protocol that combines model-based evaluation with programmatic checks and a feedback-driven optimization loop.Each generated Q\&A pair undergoes a three-tiered filtering process:

\begin{itemize}[label=\textbullet, leftmargin=8pt]
    \item \textbf{Format Checks}: This first layer of rule-based filtering standardizes the data. It includes a language check for consistency (ZH/EN) and an unexpected output check that removes conversational artifacts or other extraneous text (e.g., "Of course, I'm happy to answer your question.") from the final output.
    \item \textbf{Semantic Checks}: To ensure linguistic quality and novelty, we perform two programmatic checks. First, we filter out pairs with high perplexity ($PPL(d) > x$) to remove overly simplistic or generic content. Second, we enforce a semantic similarity threshold ($Sim(d) < x$) to eliminate duplicate or near-duplicate entries, thus ensuring data diversity.

    \item \textbf{Professional Assessment}: We employ a powerful adjudicator model to score each Q\&A pair using domain-specific metrics, including multi-link coverage, link relevance, and cooperative optimization to assess multi-stage reasoning quality, as well as practical feasibility and natural connection to evaluate real-world applicability. Pairs scoring below a predefined threshold ($Score < 8.0$) are discarded.

\end{itemize}

\textbf{Feedback Optimization}
Our pipeline operates as a closed-loop system. The insights gathered from the quality control phase are systematically fed back to refine the prompt engineering process for the generation modules. For instance, consistent detection of conversational fillers led to the inclusion of an explicit instruction to "answer the question directly," thereby improving the conciseness and data purity of subsequent generation cycles. This iterative refinement mechanism allows for the continuous improvement of both data quality and generation efficiency.

\section{Experiment}

To quantify the multi-stage reasoning capabilities of various LLMs, we conducted an extensive evaluation on the MSCoRe. We benchmarked 15 prominent LLMs, encompassing open-source models of varying scales and leading closed-source models. The performance was measured using ROUGE-L F1 scores across four domains and three distinct difficulty levels.

\subsection{LLMs PERFORMANCE COMPARISON}
The results, presented in Table~\ref{tab:llm_rouge_comparison}, reveal two primary findings. First, while the state-of-the-art closed-source model, GPT-4o (44.24 avg.), achieves the highest overall performance, the leading open-source models have become remarkably competitive. DeepSeek-7B and DeepSeek-14B outperform other major proprietary models, indicating a narrowing gap at the frontier of reasoning capabilities. Second, and more critically, we observe a universal and significant degradation in performance as task complexity increases. For example, GPT-4o's score in the Electronics domain drops from 50.21 on "Easy" tasks to 41.29 on "Hard" tasks. This consistent trend across all models underscores that true multi-stage, collaborative reasoning remains a formidable challenge, validating our benchmark's ability to effectively measure this complex capability.Furthermore, the analysis highlights a clear correlation between model scale and performance, as well as performance disparities across different industrial domains. Models generally scored higher in the Automotive and Electronics sectors compared to the more specialized Pharmaceutical domain, likely reflecting variances in their pre-training data.

\begin{table*}[htbp]
    \centering
    \caption{LLM Performance Comparison (ROUGE Scores) across Different Domains. 
        \colorbox{darkblue}{48.78} indicates the highest score in each column, 
        \colorbox{lightblue}{44.28} indicates the second highest score.
       Med. = Medium, Avg. = Average.}
    \label{tab:llm_rouge_comparison}
    \renewcommand{\arraystretch}{1.0}
    \setlength{\tabcolsep}{4pt}
    \footnotesize
    \begin{tabular}{>{\raggedright\arraybackslash}p{2.5cm}*{8}{>{\centering\arraybackslash}p{0.8cm}}*{2}{>{\centering\arraybackslash}p{1.0cm}}}
    \toprule
    \multirow{2}{2.0cm}{\centering\textbf{Model}} & \multicolumn{3}{c}{\textbf{Automotive}} & \multicolumn{3}{c}{\textbf{Pharmaceutical}} & \multicolumn{2}{c}{\textbf{Electronics}} & \multicolumn{1}{c}{\textbf{Energy}} & \multicolumn{1}{c}{\textbf{Average}} \\
    \cmidrule(lr){2-4} \cmidrule(lr){5-7} \cmidrule(lr){8-9} \cmidrule(lr){10-10} \cmidrule(lr){11-11}
    & \textbf{Easy} & \textbf{Med.} & \textbf{Hard} & \textbf{Easy} & \textbf{Med.} & \textbf{Hard} & \textbf{Easy} & \textbf{Hard} & \textbf{-} & \textbf{-} \\
    \midrule
    \multicolumn{11}{c}{\textbf{Open-Source LLMs (Small)}} \\ 
    \midrule
    Qwen2.5-1.5B & 14.76 & 12.75 & 10.75 & 10.46 & 11.51 & 11.37 & \cellcolor{lightblue}21.39 & 11.00 & 12.75 & 12.97 \\
    Llama3.2-3B & 6.34 & 4.70 & 9.19 & 9.27 & 4.76 & 4.49 & 5.63 & 4.94 & 4.70 & 6.00 \\
    Bloomz-3B & \cellcolor{lightblue}18.55 & \cellcolor{lightblue}19.47 & \cellcolor{lightblue}19.66 & \cellcolor{lightblue}16.68 & \cellcolor{lightblue}18.36 & \cellcolor{lightblue}17.83 & 11.53 & \cellcolor{lightblue}16.77 &\cellcolor{lightblue} 19.47 & \cellcolor{lightblue}17.59 \\
    Qwen2.5-3B & \cellcolor{darkblue}36.20 & \cellcolor{darkblue}30.90 & \cellcolor{darkblue}31.04 & \cellcolor{darkblue}31.37 & \cellcolor{darkblue}31.53 & \cellcolor{darkblue}30.38 &\cellcolor{darkblue}36.64 &\cellcolor{darkblue} 32.27 & \cellcolor{darkblue}30.90 &\cellcolor{darkblue} 32.36 \\
    \midrule
    \multicolumn{11}{c}{\textbf{Open-Source LLMs (Medium)}} \\ 
    \midrule
    Yi-1.5-6B & 28.81 & 23.09 & 26.72 & 22.12 & 26.82 & 25.46 & 27.85 & 24.45 & 23.09 & 25.38 \\
    Qwen2-7B & 42.37 & 34.98 & 36.07 & 33.27 & 35.10 & 33.12 & 43.15 & 35.08 & 34.98 & 36.46 \\
    Qwen2.5-7B & 31.71 & 27.64 & 29.17 & 28.99 & 28.45 & 27.51 & 33.50 & 29.14 & 27.64 & 29.31 \\
    DeepSeek-R1-7B & \cellcolor{lightblue}46.28 & \cellcolor{darkblue}40.82 & \cellcolor{darkblue}43.55 & \cellcolor{darkblue}41.81 & \cellcolor{darkblue}40.69 & \cellcolor{darkblue}39.28 & 45.71 & \cellcolor{darkblue}41.82 &\cellcolor{darkblue}40.82 & \cellcolor{darkblue}42.18 \\
    GLM4-9B & 34.18 & 25.90 & 37.74 & 26.90 & 27.12 & 26.00 & 23.79 & 26.15 & 25.90 & 28.19 \\
    Qwen2.5-14B & 30.94 & 25.59 & 32.28 & 27.03 & 27.23 & 26.10 & 31.01 & 27.59 & 25.59 & 28.15 \\
    Phi4-14B & 44.48 & 37.93 & 17.97 & 33.26 & 37.29 & 34.85 &\cellcolor{darkblue}51.94 & 34.23 & 37.93 & 36.65 \\
    DeepSeek-R1-14B & \cellcolor{darkblue}46.93 & \cellcolor{lightblue}40.78 & \cellcolor{lightblue}40.47 & \cellcolor{lightblue}40.71 & \cellcolor{lightblue}39.48 & \cellcolor{lightblue}38.29 & \cellcolor{lightblue}49.03& \cellcolor{lightblue}40.10 & \cellcolor{lightblue}40.78 &\cellcolor{lightblue} 41.84 \\
    \midrule
    \multicolumn{11}{c}{\textbf{Closed-Source LLMs}} \\
    \midrule
    Claude-3.5-haiku & 43.93 & 36.58 & 38.55 & \cellcolor{lightblue}39.21 & 35.00 & 33.71 & \cellcolor{lightblue}43.18 & 34.28 & 36.58 & 37.89 \\
    GPT-3.5-turbo & \cellcolor{lightblue}44.28 & \cellcolor{lightblue}38.38 & \cellcolor{lightblue}41.29 & 35.18 & \cellcolor{lightblue}36.73 &\cellcolor{lightblue} 34.69 & 43.13 &\cellcolor{lightblue} 35.61 & \cellcolor{lightblue}38.38 & \cellcolor{lightblue}38.63 \\
    GPT-4o & \cellcolor{darkblue}48.78 & \cellcolor{darkblue}43.21 & \cellcolor{darkblue}45.42 & \cellcolor{darkblue}43.83 & \cellcolor{darkblue}41.33 & \cellcolor{darkblue}40.92 &\cellcolor{darkblue}50.21 & \cellcolor{darkblue}41.29 & \cellcolor{darkblue}43.21 & \cellcolor{darkblue}44.24\\
    \bottomrule
    \end{tabular}

\end{table*}

\subsection{MODEL ROBUSTNESS TO TASK COMPLEXITY}
\begin{figure}[h]
\includegraphics[width=1.0\textwidth]{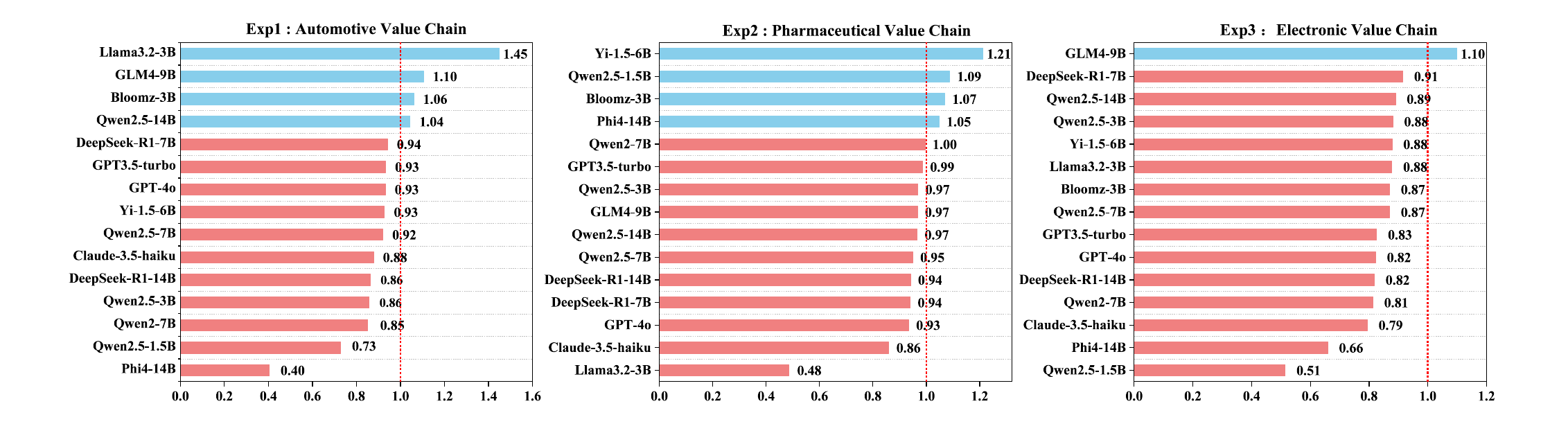} 
\caption{Model robustness evaluation across three value chain domains: performance ratio of difficult-to-easy tasks for various LLMs on Automotive (Exp1), Pharmaceutical (Exp2), and Electronic (Exp3) value chain datasets. Higher ratios (blue bars) indicate better robustness, while lower ratios (red bars) suggest performance degradation on challenging tasks.}
\label{figure_4}
\end{figure}

To move beyond absolute performance and specifically evaluate a model's stability in the face of increasing task complexity, we introduce a ratio-based robustness metric. This metric is defined as the ratio of a model's ROUGE score on \textbf{Hard} tasks to its score on \textbf{Easy} tasks within the same domain:
\begin{equation}
    \text{Robustness Ratio} = \frac{\text{ROUGE Score (Hard)}}{\text{ROUGE Score (Easy)}}
\end{equation}

 A ratio approaching 1.0 indicates exceptional robustness, signifying that the model's performance is stable and degrades minimally when transitioning from simple, single-stage problems to complex, multi-stage reasoning challenges. Conversely, a lower ratio suggests a lack of robustness or brittleness, where a model's capabilities, though potentially strong on simple tasks, falter significantly on more demanding problems. The result is shown in Figure~\ref{figure_4}.The results consistently show that leading models, such as GPT-4o, GPT-3.5-turbo, and the DeepSeek-R1 series, demonstrate high and stable robustness. Their ratios typically fall within the 0.82 to 0.99 range across all three domains. This suggests that their architectures and extensive training foster more generalizable and stable reasoning capabilities, allowing them to maintain a large fraction of their performance even as task complexity escalates.

 The metric powerfully reveals model-specific sensitivities to different domains. A striking example is Phi4-14B, which exhibits extreme brittleness in the Automotive domain with a ratio of only 0.40, the lowest in that category. However, in the Pharmaceutical domain, its ratio soars to 1.05, placing it among the most robust models. This volatility indicates that a model's reasoning stability is not an intrinsic general property but is highly contingent on its familiarity with a specific domain's relational knowledge. And we find that several models, often smaller or specialized ones like {GLM4-9B and Yi-1.5-6B, achieve a robustness ratio greater than 1.0 in certain domains. This means their ROUGE score on \textbf{Hard} tasks was actually higher than on \textbf{Easy} tasks. This phenomenon does not necessarily imply superior reasoning on harder tasks. A more plausible explanation is a stylistic artifact related to the ROUGE metric itself: \textbf{Hard} tasks often require longer, more comprehensive answers involving multiple stages. Models that are inherently more verbose may generate text that achieves higher lexical overlap on these tasks, even if the underlying logic is not perfectly sound. Therefore, this ratio also serves as an indicator of a model's behavioral tendencies and response style when faced with different problem types.

\subsection{RESULT OF FEW-SHOT-LEARNING}

\begin{table}[htbp]
\caption{Comparison of model performance under zero-shot and one-shot settings.}
\label{k-shot-learning}
\centering
\setlength{\tabcolsep}{4pt} 
\renewcommand{\arraystretch}{1.0} 
\begin{tabular}{@{}l|c|ccc}
\toprule
\textbf{Model} & \textbf{K-Shot} & \textbf{Easy} & \textbf{Med} & \textbf{Hard}  \\
\midrule
\multirow{2}{*}{Bloomz-3B} 
& 0 & 19.72 &19.64 & 19.34\\
& 1 &25.84 & 20.16 & 20.53   \\

\midrule
\multirow{2}{*}{Qwen2.5-7B} 
& 0 & 30.97 & 26.73 & 14.8  \\
& 1 & 34.42 & 29.44 & 28.78  \\

\midrule
\multirow{2}{*}{DeepSeek-14B} 
& 0 & 46.67 & 41.41 & 40.28  \\
& 1 & 42.75 & 35.69 & 31.78   \\

\midrule
\multirow{2}{*}{GPT-3.5-Turbo} 
& 0 & 43.53 & 37.75 & 38.86  \\
& 1 & 37.82 &35.68 & 33.72\\

\bottomrule
\end{tabular}

\end{table}
To investigate the impact of in-context learning on multi-stage reasoning, we conducted a comparative analysis between zero-shot (K=0) and one-shot (K=1) settings. Due to resource constraints, our analysis focused on a representative subset of four models: Bloomz-3B, Qwen2.5-7B, DeepSeek-R1-14B, and GPT-3.5-Turbo. The results are presented in the Table~\ref{k-shot-learning}. The findings reveal a striking divergence in the effectiveness of the one-shot example across different models. For Bloomz-3B and Qwen2.5-7B, the one-shot prompt provided a noticeable performance uplift. This effect was particularly pronounced for Qwen2.5-7B on \textbf{Hard} tasks, where the ROUGE score nearly doubled from 14.8 to 28.78. This suggests that for these models, the in-context example successfully demonstrates a complex reasoning pattern that is difficult to elicit in a zero-shot setting. In stark contrast, the highly capable models with strong zero-shot performance, DeepSeek-R1-14B and GPT-3.5-Turbo, exhibited a consistent and significant degradation in performance when provided with a one-shot example. For instance, DeepSeek-14B's score on \textbf{Hard} tasks dropped from 40.28 to 31.78, and a similar decline was observed for GPT-3.5-Turbo across all difficulty levels. This counter-intuitive result points to the high prompt sensitivity of these advanced models. We hypothesize that for models that already possess a robust internal strategy for these tasks (as evidenced by their high zero-shot scores), a single in-context example may conflict with their inherent reasoning process, constraining them to a less optimal solution path. This highlights a critical challenge for complex reasoning: in-context learning is not a universally guaranteed enhancement and can be detrimental to already capable models, suggesting a need for more adaptive few-shot strategies.

\subsection{"TURING TEST"}

\begin{figure}[h]
\begin{center}
    \includegraphics[width=0.5\textwidth]{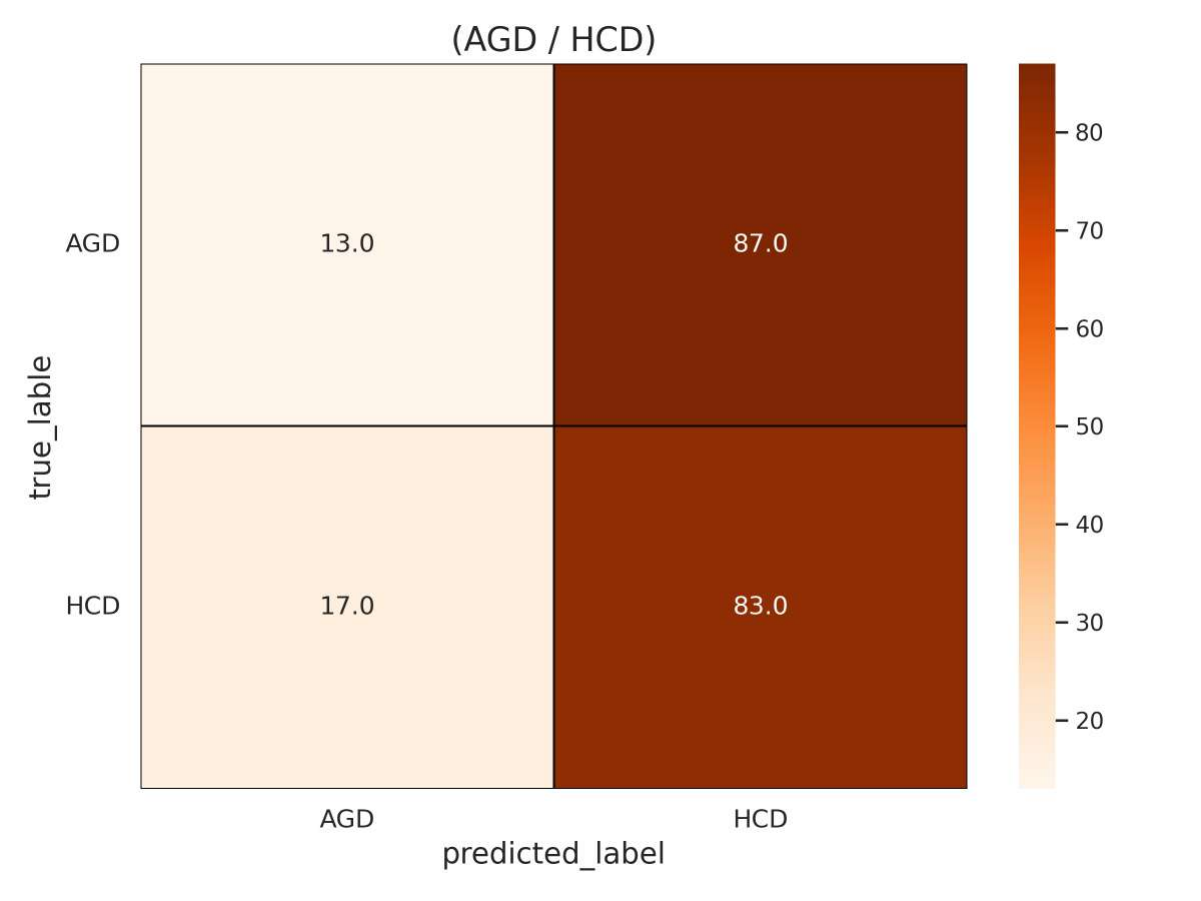} 
\caption{Confusion matrix of the Turing test comparing automatically generated data (AGD) and human-created data (HCD).}
\label{Turing-test}
\end{center}
\end{figure}
To validate the quality of our automated data generation pipeline, we designed and conducted a "Turing test" to assess whether domain experts could distinguish between data generated by our method (Automated Generated Data, AGD) and data created by humans (Human-Created Data, HCD). We presented 10 industry experts with a mixed dataset and asked them to classify each entry as either AGD or HCD. The results of this experiment are summarized in the confusion matrix shown in Figure~\ref{Turing-test}. The matrix provides a quantitative breakdown of the experts' classification performance. 

Out of 100 samples of automatically generated data (AGD), only 13.0\% were correctly identified by the experts as being generated by a large model. A significant majority, 87.0\% of the AGD samples, were incorrectly classified as human-created (HCD). This high misclassification rate strongly suggests that our generated data exhibits characteristics of human-level quality and sophistication, making it difficult for experts to distinguish from authentic human work. For the human-created data (HCD), 83.0\% of the samples were correctly identified as such. A smaller portion, 17.0\% of the HCD samples, were mistaken for AGD.

To further quantify these results, we analyze the model's performance from a classification perspective. We consider AGD as the negative class and HCD as the positive class. The overall accuracy of the experts in this test is calculated as:
\begin{equation}    
\text{Accuracy} = \frac{\text{TP} + \text{TN}}{\text{Total}} = \frac{83.0 + 13.0}{13.0 + 87.0 + 17.0 + 83.0} = \frac{96}{200} = 48.0\%
\end{equation}
This accuracy is below the 50\% threshold of random guessing, indicating that the experts were not only unable to reliably distinguish between the two data sources but performed worse than chance. This highlights the effectiveness of our generation pipeline in mimicking human-like data creation. The high rate of confusion, particularly the tendency to mistake AGD for HCD, serves as strong evidence for the quality and naturalness of the data produced by our automated system. The results of our Turing test demonstrate the exceptional quality of the datasets generated by our automated pipeline. The fact that industry experts misidentified 87\% of the AI-generated data as being human-created is a powerful testament to our method's ability to produce content that is not only coherent and relevant but also indistinguishable from that produced by human professionals. The experts' overall accuracy of 48\%, falling below random chance, further underscores the sophistication of our generated data. This experiment validates our pipeline as a highly effective and reliable method for constructing high-quality, specialized datasets, effectively closing the quality gap between automated and manual data curation.

\section{Conclusion}
We proposed MSCoRe, a novel benchmark that systematically evaluates the multi-stage collaborative reasoning abilities of LLM agents across four critical industrial value chains: automotive, pharmaceutical, electronics, and energy. Through comprehensive experiments on 15 state-of-the-art models, we demonstrated that while advanced commercial LLMs achieve the strongest overall performance, they still face substantial challenges in full-chain reasoning tasks and exhibit marked vulnerability under noisy or unstructured inputs. Furthermore, our analysis of few-shot learning revealed that in-context examples can boost weaker models but may hinder stronger ones, underscoring the complexity of prompt sensitivity in multi-stage reasoning. Finally, our expert-based Turing test validated the human-level quality of the automatically generated dataset, ensuring the reliability of MSCoRe as a robust evaluation resource. We believe that MSCoRe provides not only a rigorous testbed but also a valuable foundation for advancing the reasoning, robustness, and real-world applicability of future LLM agents in complex industrial workflows.

\bibliographystyle{unsrt}  
\bibliography{main}  

\end{document}